\pdfoutput=1  
\documentclass{article}

\usepackage{arxiv}

\usepackage[utf8]{inputenc}   
\usepackage[T1]{fontenc}      
\usepackage{booktabs}         
\usepackage{amsfonts}         
\usepackage{amsmath}
\usepackage{microtype}        
\usepackage{graphicx}
\usepackage{xcolor}
\usepackage{xspace}
\usepackage{enumitem}         
\usepackage{natbib}
\usepackage{hyperref}         
\usepackage{url}              

\definecolor{ppSampling}     {HTML}{1C5CAB}  
\definecolor{ppSamplingLight}{HTML}{5F9FE8}  
\definecolor{ppSearch}       {HTML}{AD4F18}  
\definecolor{ppSearchLight}  {HTML}{E08A4E}  
\definecolor{ppOpt}          {HTML}{1A7346}  
\definecolor{ppOptLight}     {HTML}{4FAF7D}  
\definecolor{ppInk}          {HTML}{111111}  
\definecolor{ppMuted}        {HTML}{555555}  
\definecolor{ppRule}         {HTML}{BBBBBB}  

\hypersetup{
  colorlinks=true,
  linkcolor=ppSampling,     
  citecolor=ppOpt,          
  urlcolor=ppSearch,        
  linktoc=all,
}

\usepackage[font=small,labelfont=bf,labelsep=period,justification=raggedright,
            singlelinecheck=false,skip=6pt]{caption}
\newcommand{\sysname}{\textsc{PrismPlan}\xspace}
\newcommand{\pmark}[1]{\textcolor{#1}{\rule[0.06em]{0.6em}{0.6em}}\hspace{0.4em}}

\newcommand{\repourl}{\url{https://github.com/XiaoyubotMotionTeam/PrismPlan}}

\title{\textsc{PrismPlan}: A Shared GPU Substrate for Motion Planning Across Paradigms}

\author{
  Hong Wu\thanks{Corresponding author.} \quad ChaoXuan Liu \quad ShuJie Wang \\
  XiaoYu Brain \\
  \texttt{wuhong@xiaoyubot.com} \quad \texttt{liuchaoxuan@xiaoyubot.com} \quad
  \texttt{wangshujie@xiaoyubot.com} \\
}

\renewcommand{\shorttitle}{PrismPlan: A Shared GPU Substrate for Motion Planning}

\hypersetup{
  pdftitle={PrismPlan: A Shared GPU Substrate for Motion Planning Across Paradigms},
  pdfauthor={Hong Wu, ChaoXuan Liu, ShuJie Wang},
  pdfkeywords={motion planning, GPU-parallel planning, multi-paradigm planning, cross-paradigm benchmarking, shared collision-kinematics substrate, batched collision detection},
}

\begin{document}
\maketitle

\begin{abstract}
GPU acceleration has made motion planning one to two orders of magnitude
faster\,\citep{gpucollision2012}, but it has not made the three
paradigms---sampling, search, and optimization---comparable to one another:
every GPU-accelerated planner is built around a single paradigm, so comparing
across libraries also compares their collision checkers. All three, however,
bottleneck on the same GPU primitive: batched collision detection and the batched forward
kinematics beneath it. \sysname builds that primitive once, as a shared
substrate of per-sphere collision--kinematics device code that every planner
consumes. Because they all execute the same device code, the backend is held
fixed by construction and the paradigms can be compared head-to-head. We run six
planners over 1800 MotionBenchMaker problems on three robots (7--14 DoF, one a
dual-arm whole-body composite) at an equal 2\,s budget. On the 7-DoF Panda,
sampling (pRRTC) is the most reliable paradigm at 98\%, while parallel search
(wPA$^*$SE) follows within four points at a $3\text{--}4\times$ lower median
latency, and the local optimizers trail well behind when made to solve from
scratch. On the 14-DoF dual-arm composite, sampling still solves about 90\% of
feasible problems, while no search or optimization planner clears a third of
them. Our open-source code is available at \mbox{\repourl}.
\end{abstract}

\keywords{Motion planning \and GPU-parallel planning \and Multi-paradigm planning
\and Cross-paradigm benchmarking \and Shared collision--kinematics substrate
\and Batched collision detection}

\section{Introduction}
\label{sec:intro}

Sampling, search, and optimization each win on some motion-planning problems and
lose on others\,\citep{mpinets2022}, and a practitioner with a GPU has no principled way to choose
between them: every GPU-accelerated planner is built around a single paradigm, so
any comparison across them compares collision checkers, sphere models, and
floating-point paths as much as it compares algorithms. This paper removes that
confound. We observe that all three paradigms bottleneck on one GPU
primitive---batched collision detection and the batched forward kinematics
beneath it---so we build that primitive once as a shared substrate and let each
paradigm consume it through a thin reduction. Six planners spanning the three
paradigms then run on bit-identical device code, which turns ``which paradigm,
when'' from a matter of opinion into a controlled measurement. That measurement,
over 1800 MotionBenchMaker problems on three robots at an equal budget, says that
configuration-space \emph{dimension}---not clutter---is what separates the
paradigms, and that by 14 DoF the ordering has inverted relative to what today's
GPU tooling would lead one to expect. The rest of this section sets out why the
question is worth asking, what stands in the way, and how the system answers it.

Motion planning remains a central bottleneck for robotic manipulation, and two
trends have sharpened the need for throughput rather than single-query latency
alone. First, learning-based methods (imitation, reinforcement learning, and data
generation for foundation models) consume planned trajectories by the thousands,
and the planner rather than the learner is often the rate-limiting step. Second,
reactive, contact-rich applications must replan many candidate motions
concurrently. Both regimes favor a planner that solves a batch of $B$ problems at
once and amortizes fixed costs across the batch, which maps naturally onto the
GPU's data-parallel execution model. Our substrate is batch-parallel within a
solve, but the API we ship is single-problem, so we report
intra-solve scaling (\S\ref{sec:exp-throughput}) and leave a concurrent
batch-of-$B$ entry point to future work (\S\ref{sec:batched}). The contribution
we make is orthogonal to raw throughput: it is the controlled measurement
described above, not a new speed record.

GPU-accelerated planning has responded to this need. cuRobo\,\citep{curobo2023}
and its whole-body successor\,\citep{curobov2_2026} demonstrate that
parallelized collision-free trajectory generation can be one to two orders of
magnitude faster than CPU baselines, and parallel sampling-based planners such
as pRRTC\,\citep{prrtc2025} (itself built on the vectorized collision checking
of VAMP\,\citep{vamp2024}) show comparable gains for
RRT-Connect\,\citep{rrtconnect2000}. These systems are single-paradigm artifacts
by design. A practitioner ought to be able to pick
the planner a scenario calls for, and to compare candidates on one backend;
committing to a single paradigm forecloses both. Three gaps follow.

\paragraph{Gap 1: paradigms are not exposed as first-class, comparable modules.}
Production GPU libraries are optimization-centric. Their internals may contain
geometric or graph-search components, but those act as seeding subroutines inside
the optimizer; they are not sampling-based
(RRT/PRM)\,\citep{lavalle2006,rrt1998,prm1996} or search-based
(motion-primitive lattice A$^*$/MHA$^*$/ARA$^*$)\,\citep{astar1968,mhastar2014,arastar2003}
planners that a user can invoke on their own and compare against one another. No
GPU tool today answers the question of which paradigm to use for a given problem
class.

\paragraph{Gap 2: no fair cross-paradigm comparison on a unified backend.}
Comparing planners across libraries conflates the algorithm with its collision and
kinematics implementation\,\citep{fcl2012}: differences in sphere models, ESDF
conventions, early-exit heuristics, and floating-point paths can dominate the
metrics under study. A trustworthy comparison of paradigms requires that the
collision predicate and forward kinematics be bit-identical across all planners.
No existing multi-paradigm platform provides that property, since the one
multi-paradigm framework in wide use, MoveIt\,\citep{moveit2014}, is a CPU
orchestrator whose plugins each carry their own collision representation.

\paragraph{Gap 3: whole-body high-DoF is served by optimization alone.}
On whole-body composite systems, meaning dual-arm and humanoid platforms modeled
as a single high-DoF kinematic tree with self-collision, the only GPU-native
option is trajectory optimization\,\citep{curobov2_2026}. Whether sampling or
search remains viable and competitive at these dimensionalities, on the same
backend, is unknown.

\paragraph{Key observation.}
These three gaps share a root cause. Although sampling, search, and optimization
differ in how they explore configuration space, their GPU cost is dominated by the
same primitive: batched collision detection together with the batched forward
kinematics it depends on. Sampling spends its budget validating candidate edges,
search evaluates batches of successor edges, and optimization queries costs and
gradients along a trajectory. Each of these is a batched query against a
robot-vs-scene collision--kinematics primitive, which suggests reducing planning
across paradigms to one shared GPU substrate instead of accelerating each paradigm
in isolation.

\paragraph{Approach.}
We realize this reduction as a system. A common substrate provides per-sphere
forward kinematics and sphere-vs-scene geometry, tested against oriented bounding
boxes and an in-house Euclidean signed distance field, zero-copy from the same GPU
tensors the perception and optimization stack already holds. Each paradigm then
applies its own thin \emph{reduction} on top of that substrate: a boolean
any-collision test for sampling and search, a minimum-clearance reduction for
anytime-optimal sampling, and a summed hinge penalty for optimization, which
supplies the differentiable signal a boolean cannot. The boolean collision
predicate and the forward kinematics are literally the same device code for every
planner, so a cross-paradigm benchmark holds the backend fixed by construction.

\paragraph{What is and is not novel.}
We do not claim to invent the algorithms we host: pRRTC\,\citep{prrtc2025} and its
VAMP-derived collision checking\,\citep{vamp2024}, STOMP\,\citep{stomp2011},
multi-heuristic search\,\citep{mhastar2014}, and the BIT$^*$/EIT$^*$
family\,\citep{bitstar2015,eitstar2022} all have prior origins. Nor is a
multi-paradigm framework itself new; MoveIt established that a decade ago. Our
novelty lies one layer below the framework, in the shared GPU primitive substrate
that otherwise dissimilar algorithm families consume through different call
patterns, the reduction design that makes such sharing possible, and the
comparison it enables.

\noindent Our three contributions answer the three gaps in order.
\begin{enumerate}
  \item \textbf{A unifying reduction and a shared GPU substrate.} We identify
  batched collision--kinematics as the common GPU bottleneck of all three
  paradigms and give a substrate-plus-reduction design that maps sampling to
  batched edge validation, search to batched successor/edge evaluation, and
  optimization to batched cost/gradient queries, all over one set of per-sphere
  primitives. The substrate is agnostic to the planner built on top of it: a new
  algorithm need only express its queries as one of these batched reductions, so
  the three paradigms we instantiate here are an initial population of the
  substrate and not a closed set.
  \item \textbf{Runtime-arbitrary robots and a uniform result contract.}
  We lift the compile-time fixed-robot restriction of upstream GPU planners to
  runtime-specified kinematics, and expose every planner behind one problem
  interface that returns a common result struct with wall-clock solve time. The
  uniform interface is what makes the cross-paradigm comparison, and the empirical
  conclusions we draw from it, possible. Only wall-clock time is exposed
  uniformly; a per-planner kernel/copy breakdown is not yet instrumented
  (\S\ref{sec:impl}).
  \item \textbf{A shared substrate that carries whole-body composites.} The
  substrate is parameterized only by the robot's sphere set and allowed-collision
  matrix, so composite high-DoF systems require no paradigm-specific machinery
  (\S\ref{sec:wholebody}). We evaluate this claim rather than assert it: all six
  planners run unmodified on a 14-DoF dual-arm Baxter composite, with both arms
  fused into one kinematic tree and an inter-arm self-collision specification, as
  well as on 7-DoF and 8-DoF single arms (\S\ref{sec:exp-wholebody}). The
  composite is also where the paradigms separate most sharply.
\end{enumerate}

\begin{figure}[t]
  \centering
  \includegraphics[width=0.5\linewidth]{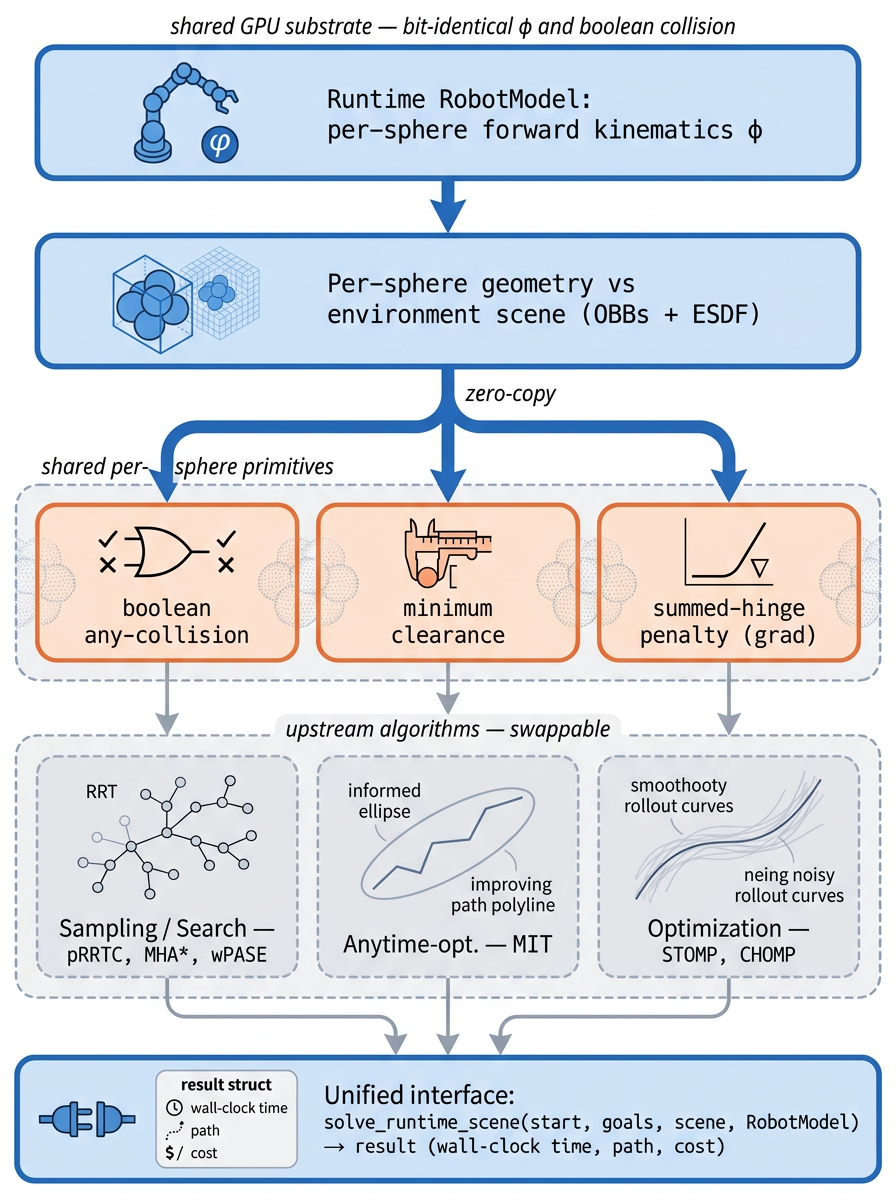}
  \caption{Architecture overview: a single batched collision--kinematics
  substrate (top) is shared, zero-copy, by sampling, search, and optimization
  modules, each applying its own reduction (middle) and reached through one
  problem interface (bottom). The forward kinematics and boolean collision
  predicate are the same device code for every planner.}
  \label{fig:arch}
\end{figure}

\noindent We call the resulting system \sysname: as a prism refracts one beam into
its component colors, it refracts a single batched collision--kinematics substrate
into a spectrum of planning paradigms, three of them in this release. The name also
reads \emph{\textbf{P}arallel \textbf{R}obot-motion \textbf{I}n a \textbf{S}hared
\textbf{M}anifold}.
Figure~\ref{fig:arch} summarizes the system: one batched
collision--kinematics substrate, consumed zero-copy by three paradigm-specific
reductions behind a single problem interface.

\section{Related Work}
\label{sec:related}

We position \sysname against four bodies of work: multi-paradigm planning
frameworks, the specific algorithms and collision engines we build upon, CPU
planning libraries, and GPU-accelerated planners. In each case the distinction
that matters is not whether multiple paradigms coexist, but what they share
underneath.

\paragraph{Multi-paradigm frameworks: the closest predecessor.}
MoveIt\,\citep{moveit2014} is the canonical multi-paradigm planning framework:
through its plugin system it unifies sampling-based planning via
OMPL\,\citep{ompl2012}, optimization via CHOMP\,\citep{chomp2009} and
STOMP\,\citep{stomp2011}, industrial motion via Pilz, and historically
search-based planning via SBPL. A unified multi-paradigm framework is not,
by itself, a contribution, and the distinction we draw is architectural.
MoveIt is a CPU orchestrator: each plugin owns its collision representation and
world model, so the planners share neither a common GPU data structure nor a
bit-identical collision predicate. MoveIt can route a query to different
paradigms, but it can neither accelerate them on a common GPU backend nor support
a controlled comparison in which the collision and kinematics implementation is
held fixed. \sysname differs here: all three paradigms consume one GPU substrate,
zero-copy, which is what delivers GPU acceleration and bit-identical comparison
at once. This is the difference between a \emph{framework} and a \emph{shared
primitive}.

\paragraph{Algorithms and collision engines we build upon.}
We credit our upstream sources explicitly and claim no part of their invention.
pRRTC\,\citep{prrtc2025} contributes GPU-parallel RRT-Connect and, following the
SIMD/SIMT vectorized collision checking of VAMP\,\citep{vamp2024}, the
edge-validation kernels we reuse. Trajectory optimization draws on
STOMP\,\citep{stomp2011}, CHOMP\,\citep{chomp2009}, and Gaussian-process motion
planning\,\citep{gpmp2_2018}; our search module
implements multi-heuristic A$^*$\,\citep{mhastar2014} and weighted Parallel
A$^*$ for Slow Expansions (wPA$^*$SE)\,\citep{pase2014,wastar1970}; and our anytime-optimal
sampler is informed by the BIT$^*$/EIT$^*$
family\,\citep{rrtstar2011,bitstar2015,eitstar2022}. Our contribution is the unification of
these paradigms over a shared GPU primitive substrate, their generalization to
runtime-specified robots, and the cross-paradigm benchmark this makes possible.

\paragraph{CPU motion planning libraries.}
Mature CPU libraries are typically organized around a single paradigm: OMPL for
sampling\,\citep{ompl2012}, search-based libraries such as SRMP\,\citep{srmp2025}
for lattice/heuristic search, and TrajOpt\,\citep{trajopt2014}/CHOMP\,\citep{chomp2009} for
optimization. They provide neither GPU batching nor a common backend across
paradigms, and so address a different regime, single-query latency, from the
high-throughput batched setting we target.

\paragraph{GPU-accelerated planning.}
cuRobo\,\citep{curobo2023} and cuRobo\,v2\,\citep{curobov2_2026} are the closest
GPU systems. They pair a highly engineered shared GPU backend (collision,
kinematics, and GPU-native perception) with trajectory optimization, and cuRobo
v2 extends this to whole-body dual-arm and humanoid systems. That backend,
however, feeds a single paradigm: geometric and graph components exist only as
seeding subroutines within the optimizer, not as independently invocable planners
that can be compared with one another. \sysname feeds one backend to three
paradigms that remain first-class and comparable, and it demonstrates sampling and
search, not only optimization, on whole-body systems. What separates us from
cuRobo\,v2 is therefore the three comparable paradigms, including on whole-body
systems, and not multi-arm support, which cuRobo\,v2 already provides.

\paragraph{Search-based and multi-robot planning.}
Conflict-based methods such as CBS\,\citep{cbs2015} and its bounded-suboptimal
variants address multi-agent coordination on the CPU. We treat multi-arm setups as
whole-body \emph{composite} systems, that is, as a single high-DoF kinematic tree
with self-collision, rather than as coordinated agents; multi-agent coordination
is left to future work.

\paragraph{Summary of the gap.}
MoveIt offers a framework but no shared GPU primitive; cuRobo offers a shared GPU
backend but serves optimization alone. To our knowledge no prior work reduces all
three planning paradigms to a single GPU primitive substrate and uses it to
benchmark them fairly, which is the gap \sysname fills.

\section{Preliminaries and Problem Formulation}
\label{sec:prelim}

\subsection{Configuration Space and Collision-Free Planning}
Let $\mathcal{C} \subseteq \mathbb{R}^{d}$ denote the configuration space of a
robot with $d$ degrees of freedom, subject to joint limits
$\mathcal{C}_{\mathrm{lim}} = \prod_{i=1}^{d}[q_i^{\min}, q_i^{\max}]$. A world
$\mathcal{W}$ contains obstacles, and the forward kinematics map
$\phi: \mathcal{C} \to SE(3)^{L}$ places the robot's $L$ links (equivalently, a
set of $n_s$ bounding spheres) in the workspace. A configuration $q$ is
\emph{free} if the robot at $q$ neither collides with $\mathcal{W}$ nor with
itself:
\begin{equation}
  \mathcal{C}_{\mathrm{free}} = \bigl\{\, q \in \mathcal{C}_{\mathrm{lim}}
  \;\bigm|\; \neg\,\mathrm{col}_{\mathrm{env}}(\phi(q),\mathcal{W}) \;\wedge\;
  \neg\,\mathrm{col}_{\mathrm{self}}(\phi(q)) \,\bigr\}.
  \label{eq:cfree}
\end{equation}
Given a start $q_{\mathrm{s}}$ and a goal set $\mathcal{Q}_{\mathrm{g}}$, a
planner seeks a continuous path
$\sigma:[0,1]\to\mathcal{C}_{\mathrm{free}}$ with $\sigma(0)=q_{\mathrm{s}}$ and
$\sigma(1)\in\mathcal{Q}_{\mathrm{g}}$, optionally minimizing a cost
$J(\sigma)$ (typically path length in $\mathcal{C}$).

The robot at $q$ is represented by $n_s$ bounding spheres
$\bigl(p_j(q),\,r_j\bigr)_{j=1}^{n_s}$ whose centers are read off $\phi(q)$, and
every geometric query in this paper is a reduction over their signed clearances
\begin{equation}
  c_j(q) \;=\; \min_{o\in\mathcal{W}} \operatorname{sd}\bigl(p_j(q),\,o\bigr) - r_j,
  \qquad
  c_{\mathrm{self}}(q) \;=\; \min_{(a,b)\in\mathcal{P}}
    \bigl\| p_a(q) - p_b(q) \bigr\| - r_a - r_b,
  \label{eq:clearance}
\end{equation}
where $\operatorname{sd}(\cdot,o)$ is the signed distance to obstacle $o$ (an
oriented bounding box, an ESDF layer, or a mesh) and $\mathcal{P}$ is the set of
sphere pairs left enabled by the allowed-collision matrix (ACM). With collision
margins $\varepsilon,\varepsilon_{\mathrm{self}}\ge 0$, the two predicates of
\eqref{eq:cfree} are thresholds on \eqref{eq:clearance}:
\begin{equation}
  \mathrm{col}_{\mathrm{env}}\bigl(\phi(q),\mathcal{W}\bigr) \Leftrightarrow
    \min_{j} c_j(q) < \varepsilon,
  \qquad
  \mathrm{col}_{\mathrm{self}}\bigl(\phi(q)\bigr) \Leftrightarrow
    c_{\mathrm{self}}(q) < \varepsilon_{\mathrm{self}}.
  \label{eq:predicates}
\end{equation}
A single per-sphere quantity $c_j$ underlies every planner in this
paper, and what separates the paradigms is only the reduction they apply to it
(Section~\ref{sec:backend}).

Operationally, almost all planning cost is incurred inside the two predicates of
\eqref{eq:cfree}. Because $\sigma$ is
validated by discretizing it into waypoints and, for each, evaluating
$\phi$ followed by the collision predicates, both the continuous feasibility test
along an edge and the pointwise test at a node reduce to batched evaluations of
$\phi$ and $\mathrm{col}$. Concretely, an edge $(q,q')$ is
validated by densifying it at a fixed resolution $\delta$ and conjoining the
pointwise predicate over the resulting waypoints,
\begin{equation}
  \mathrm{free}(q,q') \;=\;
    \bigwedge_{i=1}^{G} \neg\,\mathrm{col}\!\left(q + \tfrac{i}{G}(q'-q)\right),
  \qquad
  G \;=\; \bigl\lceil \|q'-q\| / \delta \bigr\rceil,
  \label{eq:edge}
\end{equation}
so that every edge is sampled at the same resolution regardless of its length.
Equation~\eqref{eq:edge} is the unit of work the GPU batches: the $G$ waypoints
are evaluated in parallel and the conjunction is short-circuited by a
warp-collective early exit at the first colliding waypoint. We make this the
shared substrate in Section~\ref{sec:arch}.

\subsection{Whole-Body Composites as a Single Kinematic Tree}
We model a multi-arm platform as a single \emph{composite} robot rather than as a
set of coordinating agents: the $K$ arms are concatenated into one kinematic tree
with a shared base frame, so that
\begin{equation}
  q = \bigl(q^{1},\dots,q^{K}\bigr) \in \mathbb{R}^{d},
  \quad d = \sum_{k=1}^{K} d_k,
  \qquad
  \mathcal{P} \;=\;
    \underbrace{\bigcup_{k} \mathcal{P}_{kk}}_{\text{intra-arm}}
    \;\cup\;
    \underbrace{\bigcup_{k \neq l} \mathcal{P}_{kl}}_{\text{inter-arm}},
  \label{eq:composite}
\end{equation}
where $\mathcal{P}_{kl}$ collects the ACM-enabled sphere pairs between arms $k$
and $l$. Planning then occurs in the joint high-DoF space
$\mathcal{C}\subseteq\mathbb{R}^{d}$, and the self-collision predicate
$\mathrm{col}_{\mathrm{self}}$ picks up the inter-arm terms of
\eqref{eq:composite} for free: definitions \eqref{eq:clearance} and
\eqref{eq:predicates} keep their form, and only $d$, $n_s$ and $\mathcal{P}$ grow.
Because the formulation and the substrate stay identical to the single-arm case,
all three paradigms operate on whole-body systems without paradigm-specific
machinery. Coordinated multi-agent planning, in which the arms have independent
objectives and decoupled search, is out of scope (Section~\ref{sec:related}).

\subsection{The Three Planning Paradigms}
We consider the three dominant paradigms, each with a distinct guarantee and a
distinct way of interrogating $\mathcal{C}_{\mathrm{free}}$:
\begin{itemize}
  \item \textbf{Sampling-based} (e.g.\ RRT-Connect): draws random configurations
  and connects them via local steering, which gives probabilistic completeness;
  anytime-optimal variants (BIT$^*$/EIT$^*$) additionally converge toward
  $J^{\star}$. Its dominant query is edge validation.
  \item \textbf{Search-based} (e.g.\ multi-heuristic A$^*$ over motion
  primitives): expands a discrete graph, which gives completeness with bounded
  suboptimality $w\!\cdot\!J^{\star}$. Its dominant query is successor-edge
  evaluation.
  \item \textbf{Optimization-based} (e.g.\ STOMP, CHOMP): iteratively refines a
  full trajectory against a smooth cost, which gives local optimality and soft
  constraint handling. Its dominant query is a cost or gradient evaluation along
  the trajectory.
\end{itemize}
The guarantees differ, but each dominant query is a batched interrogation of the
same $\phi$-then-$\mathrm{col}$ pipeline, and all that differs is the
\emph{reduction} applied to the per-sphere clearances of \eqref{eq:clearance}
(Eqs.~\eqref{eq:red-bool}--\eqref{eq:red-hinge}, Section~\ref{sec:arch}).

\subsection{Batched Evaluation and Intra-Solve Parallelism}
\label{sec:batched}
The GPU's data-parallel model exposes parallelism at two levels: across
independent problems, and within a single solve, over the many sphere and waypoint
evaluations of each collision query. Our substrate is batch-parallel at the second
level. The sampling planner, for instance, launches $W$ parallel tree-extension
blocks per solve, four threads to an edge at granularity $g$, and the search module flushes an
entire independent set of candidate edges to one batched kernel. We therefore
characterize intra-solve scaling, that is, how solve time and device memory move
with parallel width $W$ and tree capacity (\S\ref{sec:exp-throughput}). A
concurrent batch-of-$B$ entry point that amortized scene upload and kinematic
tables across many simultaneous queries would be a natural extension of this
design, but it is not part of the current single-problem API, so we treat it as
future work instead of reporting throughput-vs-$B$ curves the shipped code cannot
produce. Timing is reported as wall-clock solve time (\S\ref{sec:iface}); a
kernel/copy decomposition is not yet instrumented uniformly across planners.

\section{System Architecture}
\label{sec:arch}

The architecture realizes a single idea: one batched collision--kinematics
substrate, consumed by three paradigm-specific reductions.
Section~\ref{sec:backend} describes the substrate and the reductions,
Sections~\ref{sec:sampling}--\ref{sec:opt} the three paradigm modules built on it,
Section~\ref{sec:wholebody} whole-body operation, and Section~\ref{sec:iface} the
unified interface that makes cross-paradigm comparison well posed.

\subsection{The Shared GPU Substrate}
\label{sec:backend}

\paragraph{Per-sphere geometry substrate.}
The substrate is a set of device primitives that transform a world-space sphere
into each obstacle's local frame and evaluate its collision or signed clearance
against (i) oriented bounding boxes, (ii) an in-house Euclidean signed distance
field (ESDF)\,\citep{voxblox2017}, and (iii) triangle meshes. They follow the
sign and indexing conventions of the perception stack, so their results match the
optimizer's own world model. On top of these, a cooperative check evaluates the
full robot---all $n_s$ of its bounding spheres---in two phases: a conservative
approximate sphere set is tested first and, being an outer bound, admits a
warp-collective early exit whenever it is already clear; only when that set
reports a possible collision is the detailed sphere set evaluated to confirm.
Writing $\hat c_j$ for the clearance of the approximate set, the enclosure
property gives a one-sided bound, so the two-phase test is exact rather than
approximate:
\begin{equation}
  \hat c_j(q) \le c_j(q) \;\; \forall j
  \qquad \Longrightarrow \qquad
  \min_j \hat c_j(q) \ge \varepsilon \;\Rightarrow\; \min_j c_j(q) \ge \varepsilon.
  \label{eq:twophase}
\end{equation}
An approximate miss is conclusive and skips the detailed pass, while an
approximate hit means only ``maybe'' and must be confirmed at full resolution
before a configuration is rejected. Forward kinematics comes from a runtime
routine that consumes a data-driven robot model rather than a
compile-time-fixed robot (Section~\ref{sec:iface}).

\begin{figure}[t]
  \centering
  \includegraphics[width=0.72\linewidth]{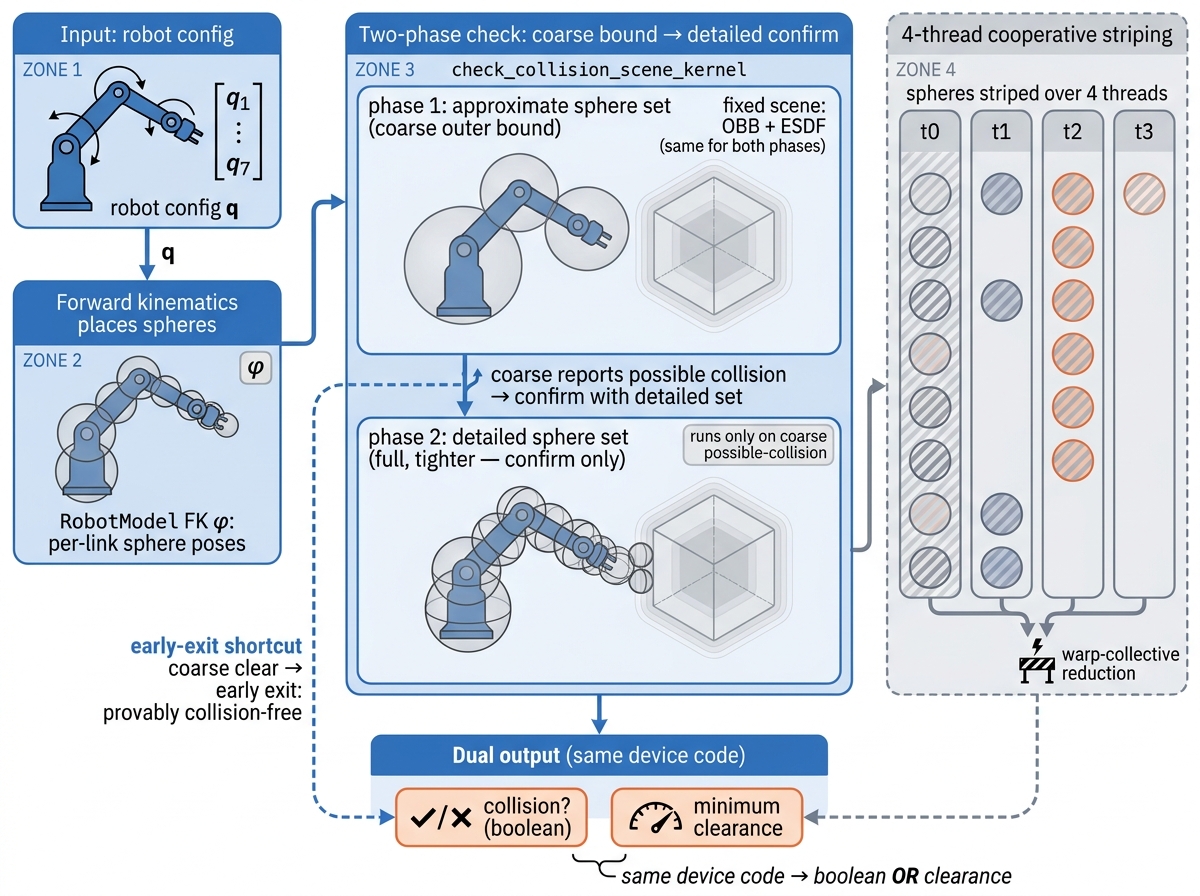}
  \caption{Shared per-sphere geometry substrate and its cooperative evaluation:
  runtime FK places the robot's spheres, a two-phase approximate$\to$detailed
  check with a warp-collective early exit evaluates them against the scene, and
  the $n_s$ spheres are striped over 4 cooperating threads. The same device code yields either
  a boolean verdict or a minimum clearance.}
  \label{fig:substrate}
\end{figure}

\paragraph{Three paradigm-specific reductions.}
The three paradigms differ only in how they reduce the per-sphere clearances of
\eqref{eq:clearance} into the signal their algorithm consumes
(Table~\ref{tab:reductions}):
\begin{align}
  \rho_{\mathrm{bool}}(q)  &= \Bigl[\, \min_{j} c_j(q) < \varepsilon \,\Bigr]
    \;\vee\; \bigl[\, c_{\mathrm{self}}(q) < \varepsilon_{\mathrm{self}} \,\bigr],
    \label{eq:red-bool} \\[2pt]
  \rho_{\mathrm{clr}}(q)   &= \min_{j} c_j(q),
    \label{eq:red-clr} \\[2pt]
  \rho_{\mathrm{hinge}}(q) &= w_c\Bigl[\, \sum_{j=1}^{n_s}
      \max\bigl(0,\ \varepsilon_{w} - c_j(q)\bigr)
    \;+\; \max\bigl(0,\ \varepsilon_{s} - c_{\mathrm{self}}(q)\bigr) \Bigr].
    \label{eq:red-hinge}
\end{align}
Sampling and search consume the boolean any-collision verdict
\eqref{eq:red-bool}, which is identical to $\mathrm{col}$ in
\eqref{eq:predicates}. Anytime-optimal sampling additionally consumes the minimum
clearance \eqref{eq:red-clr} to order and prune edges, which enters an edge cost
$\|q'-q\|\bigl(1+\min(\alpha/\rho_{\mathrm{clr}},\,\bar\alpha)\bigr)$ with
$\alpha$ the clearance weight ($\alpha=0$, i.e.\ pure $\mathcal{C}$-space length,
in every reported run). Optimization consumes the differentiable penalty
\eqref{eq:red-hinge}, a sum of per-sphere hinges plus one hinge on the worst
self-collision pair, over activation bands
$\varepsilon_{w},\varepsilon_{s}$ that may be wider than the verdict margin
$\varepsilon$, scaled by a collision weight $w_c$. Only $\rho_{\mathrm{hinge}}$
is piecewise differentiable and non-flat outside contact, which is why an
optimizer cannot be driven by $\rho_{\mathrm{bool}}$. All three reductions read the
same $c_j(q)$, produced by the same FK and the same substrate primitives, so the
underlying geometry is identical across paradigms by construction.

\begin{table}[t]
  \centering
  \caption{The three paradigms share one substrate and differ only in the
  reduction applied to the per-sphere collision/clearance result
  (Eqs.~\eqref{eq:red-bool}--\eqref{eq:red-hinge}). Self-collision
  is evaluated by the same per-sphere machinery, gated by the robot's
  allowed-collision matrix (ACM). Kernel names are the shipped implementations.}
  \label{tab:reductions}
  \small
  \begin{tabular}{@{}lllll@{}}
    \toprule
    Paradigm & Consumed signal & Reduction (kernel) & Self-collision & Guarantee \\
    \midrule
    Sampling / Search & valid / invalid & boolean any-collision & ACM-gated & prob.\ complete / \\
      & & & sphere pairs & bounded subopt. \\
    Anytime-opt.\ (sampling) & edge cost + clearance & minimum clearance & ACM-gated & asymptotically \\
      & & & & optimal \\
    Optimization & smooth cost + gradient & summed hinge penalty & ACM-gated & locally optimal \\
    \bottomrule
  \end{tabular}
\end{table}

\paragraph{Bit-identical anchor.}
The boolean verdict is a single device routine, the two-phase
checker of \eqref{eq:twophase}, and every planner calls it: the
anytime sampler directly, the two search planners through the batched
edge evaluator, the sampling planner from its
edge-validation inner loop, and the two optimizers from the validity kernel that
decides whether the trajectory they return is collision-free. The optimizers
deserve a word of precision. Their cost cannot be the boolean, since CHOMP needs a
gradient and a boolean has none, while STOMP needs cost differences between
rollouts that a boolean would collapse; so they consume the summed-hinge
reduction while optimizing. Their verdict, meaning the test that decides whether a
solve counts as solved, is the shared routine. Two paths bypass it, and no number
we report exercises either: the mesh-collision branch, for which the anchor has no
counterpart and which the OBB-based MotionBenchMaker evaluation never takes, and a
legacy entry point predating the runtime scene representation. Since all three
paradigms obtain their boolean feasibility verdict from one device routine and
their link poses from one forward kinematics, the feasibility test is
bit-identical across planners. That is the technical precondition for the
comparison in Section~\ref{sec:xparadigm}, where any measured difference must
reflect the algorithm rather than its collision or kinematics implementation.
Determinism is aided by a fixed
sampler seed (the sampling planners draw from a Halton sequence initialized once
per solve) and by the shared fast-math forward kinematics
(\S\ref{sec:impl}); we do not claim bit-for-bit run-to-run identity under the
non-deterministic atomics used in tree insertion, only that the feasibility
predicate itself is the same device code for all planners.

\paragraph{Zero-copy scene and memory layout.}
The scene (OBBs, ESDF, and meshes) is held in GPU tensors shared zero-copy with
the perception and optimization stack, which avoids a redundant world
representation and the copies it would entail. Sphere positions and per-sphere
clearances are kept in structure-of-arrays layout so that the 4 cooperating
threads read contiguous lanes, and the scene buffers persist on the device across
solves, so only the per-problem start and goal and, when the world changes, the
OBB set are uploaded.

\subsection{Sampling-Based Module: \texorpdfstring{pRRTC and MIT$^*$}{pRRTC and MIT*}}
\label{sec:sampling}
The sampling module provides three variants, all of which take feasibility and
clearance from the substrate of \S\ref{sec:backend}: GPU-parallel
RRT-Connect with SIMT batched edge validation, a constrained variant (cpRRTC)
that projects sampled configurations onto a task manifold before
validation\,\citep{cbirrt2009,constrained2018}, and an anytime-optimal sampler
with reverse search. Following pRRTC\,\citep{prrtc2025}, RRT-Connect grows
two trees, one from the start and one from the goal, extending them concurrently
across $W$ parallel blocks per iteration and periodically attempting to connect
them; each candidate edge is validated by discretizing it into $g$ waypoints and
issuing the batched collision check, with a warp-collective early exit on the
first colliding waypoint. The anytime-optimal sampler
(MIT$^*$\,\citep{mitstar2025}) maintains forward and reverse trees with a batched
cost-propagation and prune cascade, and records
a cost/time history from which an anytime curve can be read off.

\subsection{Search-Based Module: \texorpdfstring{MHA$^*$ and wPA$^*$SE}{MHA* and wPA*SE}}
\label{sec:search}
The search module implements two planners over a discretized action set:
multi-heuristic A$^*$ (MHA$^*$) and weighted Parallel A$^*$ for Slow Expansions
(wPA$^*$SE). Both are CPU/GPU hybrids and not pure-GPU searches: the priority-queue
search runs on the host, while edge feasibility and cost are evaluated lazily in
GPU batches through the same substrate. The host search defers candidate edges,
flushes them to the GPU in batches, and consumes the returned \{valid, cost,
clearance\} triples. wPA$^*$SE additionally selects, each round, a set of mutually
independent OPEN states and flushes all of their successor edges in one batch,
which maps the reference algorithm's ``parallel slow expansions'' onto GPU batch
width instead of CPU threads and retains a bounded suboptimality of $w$. Because the same wPA$^*$SE
also exists as a CPU planner in SRMP\,\citep{srmp2025}, it affords a direct
same-algorithm, different-substrate comparison (\S\ref{sec:xparadigm}). The
batching policy is deferral: the host search accumulates candidate edges
until it holds either the successors of one expansion (MHA$^*$) or those of a whole independent set
(wPA$^*$SE), then flushes them in a single batched launch
sized to the number of pending edges; larger independent sets give wider GPU
batches at the cost of expanding less greedily.

MHA$^*$ runs a small fixed set of inadmissible heuristics
$h_1,\dots,h_N$ behind one consistent anchor $h_0$, keying queue $i$ by
$\mathrm{key}_i$ and expanding it only while its minimum key stays within an
anchor bound:
\begin{equation}
  \mathrm{key}_i(s) = g(s) + w_1\,h_i(s),
  \qquad
  \min_{s\in\mathrm{OPEN}_i} \mathrm{key}_i(s)
    \;\le\; w_2 \min_{s\in\mathrm{OPEN}_0} \mathrm{key}_0(s),
  \label{eq:mha-key}
\end{equation}
which bounds the returned solution by $w_1 w_2 \cdot J^{\star}$. The
inadmissible heuristics are per-DoF reweightings of the same anchor,
\begin{equation}
  h_i(q) = \min_{g\in\mathcal{Q}_{\mathrm{g}}} \bigl\| W_i\,(q-g) \bigr\|_2,
  \qquad
  W_i = \mathrm{diag}(w_{i,1},\dots,w_{i,d}), \quad w_{i,n} \in [1,\eta],
  \label{eq:mha-heur}
\end{equation}
with odd $i$ inflating the proximal joints and even $i$ the distal ones. The
constraint $w_{i,n}\ge 1$ is what makes them genuinely inadmissible
($h_i \ge h_0$ pointwise), which is the point of an inadmissible queue: it is
allowed to overestimate in order to escape a local minimum faster, while
\eqref{eq:mha-key} retains the guarantee. Weights below $1$ would leave every
``inadmissible'' queue dominated by the anchor, so that MHA$^*$ would spend most
of its budget on a search less informed than plain weighted A$^*$. That failure
mode is invisible in the structure of the algorithm and visible only in
\eqref{eq:mha-heur}. On a branching whole-body robot the reweighting splits along
the branches at no extra cost: Baxter's DoF $0$--$6$ are one arm and $7$--$13$
the other.

wPA$^*$SE keys a single OPEN queue by $f$ and, each round, expands a set
$\mathcal{I}$ of mutually independent states,
\begin{equation}
  f(s) = g(s) + w\,h(s),
  \qquad
  \mathcal{I} \ \text{s.t.}\ \forall\, s,s'\in\mathcal{I}:\;
    g(s) \le g(s') + w\,h(s'\!\to\!s),
  \label{eq:pase}
\end{equation}
under the pairwise heuristic $h(s'\!\to\!s)$; the successor edges of all of
$\mathcal{I}$ are flushed to the GPU in one batch, and $|\mathcal{I}|$ is capped
at the configured parallelism. We do not claim a
GPU-resident frontier: the priority queues and the independent-set selection run
on the host, and only edge evaluation is on the GPU.

\subsection{Optimization-Based Module: STOMP and CHOMP}
\label{sec:opt}
The optimization module provides two representatives that share the substrate
but differ in how they use it: a derivative-free optimizer (STOMP), which rolls
out noisy trajectories in batch and reweights them by cost, and a gradient-based
optimizer (CHOMP), whose covariant update follows an obstacle gradient obtained
as the finite-difference derivative of the same per-timestep cost STOMP
evaluates. The world-collision cost is the summed-hinge reduction
$\rho_{\mathrm{hinge}}$ of \eqref{eq:red-hinge}, giving a smooth per-timestep penalty for both ESDF and
OBB obstacles. Taking CHOMP's gradient as a finite difference of that identical
cost keeps the gradient-based representative on the shared substrate; a
separately derived ESDF gradient could drift away from the anchor.

Both optimizers represent a trajectory as $\theta\in\mathbb{R}^{d\times T}$ with
pinned endpoints, and both measure smoothness with the same five-point
acceleration matrix $R$. STOMP draws $K$ noisy rollouts
$\theta_k = \theta + \epsilon_k$ with $\epsilon_k\sim\mathcal{N}(0,R^{-1})$
(via a Cholesky factor of $R^{-1}$), scores each
rollout per dimension $n$ and timestep $t$ against the substrate, and turns the
scores into importance weights:
\begin{align}
  S_k(n,t) &= \rho_{\mathrm{hinge}}\bigl(\theta_k(\cdot,t)\bigr)
    \;+\; \lambda_c\, \theta_k(n,t)\,\bigl[R\,\theta_k(n,\cdot)\bigr](t),
    \label{eq:stomp-cost}\\
  P_k(n,t) &= \frac{\exp\bigl(-h\,\tilde S_k(n,t)\bigr)}
                   {\sum_{k'=1}^{K}\exp\bigl(-h\,\tilde S_{k'}(n,t)\bigr)},
    \qquad
    \tilde S_k = \frac{S_k - \min_{k'} S_{k'}}{\max_{k'} S_{k'} - \min_{k'} S_{k'}},
    \label{eq:stomp-prob}
\end{align}
where $\lambda_c$ weights the control cost, $h$ is the cost sensitivity, and the
normalization is taken over the $K$ rollouts at that $(n,t)$. The weighted noise average is then projected back onto
the smooth subspace the noise was drawn from before it is applied:
\begin{equation}
  \delta\theta(n,t) = \sum_{k=1}^{K} P_k(n,t)\,\epsilon_k(n,t),
  \qquad
  \theta \leftarrow \theta + M\,\delta\theta,
  \quad M \propto R^{-1},
  \label{eq:stomp-update}
\end{equation}
which is necessary because the per-timestep weights of \eqref{eq:stomp-prob}
vary independently and $\delta\theta$ is not smooth on its own. CHOMP
instead takes a covariant gradient step preconditioned by that same metric
($A = R + 10^{-8}I$), with its obstacle term a central finite difference of the
identical per-timestep cost:
\begin{equation}
  \theta \leftarrow \theta - \eta\,A^{-1}
    \bigl(\nabla_{\mathrm{obs}}(\theta) + \lambda\,R\,\theta\bigr),
  \qquad
  \bigl[\nabla_{\mathrm{obs}}\bigr]_{n,t} =
    \frac{\rho_{\mathrm{hinge}}\bigl(\theta(\cdot,t) + \epsilon_{\mathrm{fd}} e_n\bigr)
        - \rho_{\mathrm{hinge}}\bigl(\theta(\cdot,t) - \epsilon_{\mathrm{fd}} e_n\bigr)}
         {2\,\epsilon_{\mathrm{fd}}}.
  \label{eq:chomp-update}
\end{equation}
Here $\eta$ is the step size, $\lambda$ the smoothness weight,
$\epsilon_{\mathrm{fd}}$ the finite-difference step, and $e_n$ the $n$-th
coordinate direction.
The $\rho_{\mathrm{hinge}}$ of \eqref{eq:chomp-update} is the
$\rho_{\mathrm{hinge}}$ of \eqref{eq:stomp-cost}: the same
Equation~\eqref{eq:red-hinge}, evaluated by the same device code. The
derivative-free and gradient-based representatives thus differ in how they use
the substrate, not in what they measure. Both follow their published
formulations\,\citep{stomp2011,chomp2009}.

\subsection{Extension to Whole-Body Composites}
\label{sec:wholebody}
Because the substrate is parameterized by the robot model's sphere set
and ACM, whole-body operation requires no new machinery: the composite kinematic
tree of Section~\ref{sec:prelim} presents a larger $d$, more spheres, and
inter-arm self-collision pairs, and nothing else changes. All three paradigms
run on whole-body systems through the identical substrate, with no
paradigm-specific code path. The released implementation instantiates such a
composite: the Baxter model concatenates both 7-DoF arms into a
single 14-DoF tree of 15 joint frames, 75 spheres, and 80 inter-arm
self-collision sphere-range pairs, and is loaded through the same runtime path as
a single arm. Section~\ref{sec:exp-wholebody} measures all six planners on it and
confirms that the architectural property holds: no planner needed modification.
Their empirical behavior, however, diverges sharply at 14 DoF.

\subsection{The Unified Planner Interface}
\label{sec:iface}
Every planner is exposed behind a single entry point that takes the same problem
description---start, goal set, and a shared scene handle---plus a robot model
loaded at runtime, and returns a result struct with a consistent timing
contract, namely a wall-clock solve time measured the same way for every
paradigm. Identical inputs, an identical backend, and comparable outputs are what
make the cross-paradigm benchmark well posed. The result carries a success flag,
the geometric path, a scalar cost, the wall-clock solve time, and
planner-specific counters (tree size for sampling; expansions and
edges-evaluated for search). The timing field
is uniform across planners; a finer kernel/copy split is not currently populated
(\S\ref{sec:impl}).

\section{Implementation}
\label{sec:impl}

\paragraph{Languages and framework.}
The shared substrate and all three planner families are implemented in
CUDA/C++: the per-sphere geometry primitives, runtime kinematics, and the
paradigm-specific reduction kernels of Section~\ref{sec:backend} are compiled
device code, while tree/graph bookkeeping (the sampling trees, the CPU-side
MHA$^*$ open lists, and STOMP's outer loop) is host C++. A single Python module
exposes every planner's entry point, so the same backend is reachable from C++ callers and Python scripts
without a second collision or kinematics implementation. The implementation is
roughly 17.5k lines of CUDA/C++ (device kernels, runtime kinematics, and the three
planner families) and 2.5k lines of Python (the plugin layer, benchmark harness,
and substrate bindings). It targets CUDA 12.x and is built for compute capability
8.6 (the RTX~3060 used in \S\ref{sec:exp}); no compute-capability-specific
intrinsics beyond the fast-math FK are used.

\paragraph{Runtime robot model.}
Where the upstream pRRTC fixes the robot at compile time and supports a small
set of built-in arms, \sysname loads kinematics and collision geometry into a
runtime robot model. It carries per-joint fixed transforms and
joint types, from which the runtime FK is evaluated; per-link sphere sets for the
collision substrate; and an allowed-collision matrix stored as the set of sphere
pairs exempt from self-collision. A single binary can plan for an
arbitrary URDF and, in the whole-body case, for a composite multi-arm tree
assembled at load time. This release ships three such models, all consumed by the same binary: the 7-DoF Panda (8 joint frames
including the fixed base, 59 spheres), the 8-DoF Fetch (9 frames, 111 spheres,
including a prismatic torso joint), and the 14-DoF dual-arm Baxter composite (15
frames, 75 spheres).

\paragraph{Integration.}
The library is organized as a set of planner plugins behind one common planner
and pipeline abstraction, so a caller selects a paradigm without touching the
shared substrate. Scenes are supplied as
OBB primitives (and, where available, an in-house ESDF) held in GPU tensors that the
substrate reads in place, so perception output can feed the planners without a
host round trip. We call the abstraction integration-ready only in a limited
sense: the present release is exercised through its Python API and the harness of
\S\ref{sec:exp-setup}, and ships a ROS\,2 node that exposes every planner through
a native, MoveIt-free planning service and ingests collision geometry from a
scene topic (non-box primitives are over-approximated to enclosing OBBs, matching
the substrate's collision path).
The ROS\,2 adapter is provided as code but not exercised in our evaluation;
simulator (MuJoCo/PyBullet/Sapien/Genesis) adapters remain a planned integration
surface.

\paragraph{Natural-language front end.}
The Python surface also supports a natural-language task-specification front end
for LLM-agent callers. This is an integration convenience, not a core
contribution, and is not evaluated.

\paragraph{Determinism and reproducibility.}
Fair cross-paradigm comparison requires that the shared backend be numerically
reproducible. All planners use the same fast-math trigonometric intrinsics in the forward
kinematics and the same ESDF voxel indexing with a \emph{robust-floor}
rounding convention, so that grid-cell assignment is bit-stable at voxel
boundaries and identical across all planners. The sampling planners seed a Halton sequence once per solve, giving
run-to-run stable sample streams; we do not serialize the non-deterministic
atomics used for parallel tree insertion, so exact node orderings can vary while
the feasibility verdict does not.

\section{Experimental Evaluation}
\label{sec:exp}

Our evaluation is organized around one question per subsection. We first fix the
setup (\S\ref{sec:exp-setup}), then report throughput scaling
(\S\ref{sec:exp-throughput}), single-arm and whole-body comparisons
(\S\ref{sec:exp-singlearm}, \S\ref{sec:exp-wholebody}), the cross-paradigm study
that is only possible on a shared backend (\S\ref{sec:xparadigm}), ablations of
the backend itself (\S\ref{sec:exp-ablation}), and an integration demonstration
(\S\ref{sec:exp-integration}).

\subsection{Experimental Setup and Protocol}
\label{sec:exp-setup}
\paragraph{Hardware and software.}
All experiments run on a single NVIDIA RTX~3060 (12\,GiB, compute capability
8.6, driver 570.211.01, CUDA 12.6), single-GPU throughout. The compiled
planner module is driven from Python; solve times are
measured inside the planner (wall-clock). The test machine hosts a
co-resident GPU process, so absolute device-memory figures are reported as a
delta above a per-measurement baseline to isolate each planner's own allocation
(\S\ref{sec:exp-throughput}).

\paragraph{Controlling host load.}
Two of the six planners run their search on the CPU and offload only edge
evaluation (\S\ref{sec:search}), so under a fixed wall-clock budget any
competing host load costs them expansions and therefore success. We have seen
this happen: an earlier attempt at the run reported here was contaminated by an
unrelated desktop application consuming roughly 1.6 cores, which cost wPA$^*$SE
13 problems and halved MHA$^*$'s success on the cluttered \texttt{cage} family.
The drift was large enough to be mistaken for an algorithmic effect and left no
trace in the output. All reported runs are taken on an otherwise idle machine, with host load
checked before launch and sampled every 30\,s throughout; the
reported run held a median load average of 1.33 with our own process the busiest
in 458 of 463 samples. We record this because a CPU/GPU hybrid planner
benchmarked against a wall clock is only as reproducible as the host it runs on,
and that is one confound the shared substrate does not remove.

\paragraph{Baselines and scope of the evaluation.}
The evaluation in this release is deliberately internal. Its purpose is the
cross-paradigm study that the shared backend uniquely enables, so the primary
measurement is among the six planners driven through one interface on a
bit-identical backend (\S\ref{sec:xparadigm}), alongside intra-solve scaling and a
memory ceiling for the sampling planner (\S\ref{sec:exp-throughput}). A
systematic external-baseline study is the natural next step and is not run here.
Where a claim turns on whether a result is a property of the algorithm or of our
port, we cite the corresponding CPU reference at that point
(\S\ref{sec:exp-wholebody}). What we do not provide is a like-for-like
performance table against the systems below:
\begin{itemize}[leftmargin=1.4em,itemsep=1pt,topsep=2pt]
  \item \textbf{OMPL}~\citep{ompl2012} (CPU sampling via MoveIt) and
        \textbf{SRMP}~\citep{srmp2025} (CPU search, which ships the same wPA$^*$SE
        we run on the GPU) would isolate the substrate's contribution for a fixed
        algorithm; wiring the MoveIt/CPU adapters for these is future work.
  \item \textbf{Upstream pRRTC}~\citep{prrtc2025} --- what we add over it is
        runtime-arbitrary robots on a shared, comparable backend rather than raw
        speed; we host it as our sampling planner instead of benchmarking against
        a separate build.
\end{itemize}
We also do not treat cuRobo/cuRobo\,v2~\citep{curobo2023,
curobov2_2026} as a baseline: it exposes only the optimization paradigm as a
first-class planner, so a head-to-head number would compare implementations
rather than paradigms, which is orthogonal to our shared-substrate,
cross-paradigm question. We discuss it as the closest related system in
Section~\ref{sec:related}.

\paragraph{Metrics.}
We report success rate; per-solve wall-clock time (the median taken over solved
problems only, so that a planner which burns the budget on failures is not
credited with a fast median); path cost (configuration-space length for the
sampling, search, and anytime planners; STOMP and CHOMP report their own internal
objective, so cost is not commensurable across paradigms and we compare it only
within one); and, for the sampling planner, device-memory footprint and
intra-solve time scaling with parallel width and tree capacity. A
wall/kernel/copy decomposition is not instrumented in the current release, so we
report wall-clock time only.

\paragraph{Scenes.}
We evaluate on three MotionBenchMaker~\citep{motionbenchmaker2021} robots of
increasing dimension, 1800 problems in total:
\begin{itemize}[leftmargin=1.4em,itemsep=1pt,topsep=2pt]
  \item \textbf{Panda} (7 DoF, 59 spheres) --- the problem set as distributed by
        \texttt{robometrics}: 800 problems across 8 scene families (bookshelf
        small/tall/thin, box, box-flipped, cage, table-pick, table-under-pick),
        100 each, all with joint-space goals.
  \item \textbf{Fetch} (8 DoF including the prismatic torso, 111 spheres) ---
        700 problems across the 7 scene families the upstream MotionBenchMaker
        distribution ships for this robot, 100 each.
  \item \textbf{Baxter} (14 DoF, 75 spheres) --- the dual-arm whole-body
        composite of \S\ref{sec:wholebody}: both arms fused into one kinematic
        tree with an inter-arm self-collision specification (80 sphere-range
        pairs). 300 problems, 100 each from the three
        \texttt{bookshelf\_tall\_both\_arms} difficulty tiers.
\end{itemize}
A problem whose goal configuration is already in collision under our OBB
inflation cannot be solved by any planner; the harness flags these identically
for every planner (15 of 800 Panda, 22 of 700 Fetch, 28 of 300 Baxter), and we
verified that all six agree on exactly which problems they are. We
report the raw \texttt{successes/problems} rate for comparability with the MBM
literature and give the feasible-subset rate alongside it where the distinction
changes the reading. The intra-solve scaling and memory sweeps
(\S\ref{sec:exp-throughput}) use a fixed 24-problem Panda subset (box,
table-pick, cage; 8 each) held constant across all sweep points.

\paragraph{Protocol.}
Every problem is solved once per planner with a \textbf{2\,s} per-problem time
budget; all planners see the same start/goal pairs and the same OBB scene, and
query the same bit-identical collision predicate. A single OBB inflation is used
throughout, and the pedestal beneath the robot base is excluded from collision
(it is a mount, not an obstacle). We report success rate, median
solve time over solved problems, and per-scene breakdowns. A stricter
SRMP-matched protocol---goal perturbation of $\pm2^\circ$ with 10$\times$
(single-arm) / 5$\times$ (whole-body) repetitions and a coefficient-of-variation
consistency metric, at 5\,s/20\,s/10\,s budgets---is the intended protocol for the
external-baseline comparison and is left as future work; the numbers here are
single-trial and should be read as such. Two noise floors bound how finely the
success rates may be read. Run-to-run jitter at a fixed wall-clock budget is
roughly $\pm1$ problem, and the binomial sampling error at 95\,\% confidence ranges
from $\pm$0.9 points at a 98\,\% success rate to $\pm$2.9 points at 77\,\% on the
800-problem Panda set; on the 300-problem Baxter set, where the smaller
denominator dominates, it reaches $\pm$5.5 points. Differences of a
point or two between planners are not resolvable, and we flag in the
text where a gap is close to that floor rather than reporting rankings the data
cannot support.

\paragraph{Auditing the equal budget.}
An equal \emph{nominal} time budget is not an equal \emph{effective} one. A
planner whose solve loop can terminate for a non-time reason will silently
return early and be scored as if it had used the budget the others consumed, so
before producing any number in this section we audited every planner's solve
loop for termination conditions other than the clock. This was not a formality:
four of the six had one. STOMP and CHOMP carried iteration caps
of 20 and 100 that ended their loops after roughly
1\,\% of the intended budget; MIT$^*$'s GPU adjacency buffer
of 50\,000 nodes ended its search at a 1211\,ms median under a
nominal 2\,s budget. Both classes of cap are now lifted so that the clock is the
binding constraint: MIT$^*$'s median moves to 2009\,ms, genuinely time-bound, at a
cost of 25.6\,MB of device memory for the larger graph and no change in its
success rate or path cost. pRRTC's sample and iteration ceilings were measured
not to bind at 2\,s (identical success at $10^6$ and $4\!\times\!10^6$), and
MHA$^*$ and wPA$^*$SE terminate only on the clock or an exhausted queue. We report
this because the failure mode is invisible in the output, where the affected
planners merely look fast and weak, and because it invalidates any
cross-paradigm comparison built on top of it.

A control run isolates how much of this matters. Holding the code fixed and
halving the budget back to 1\,s changes the Panda results by at most 4.0
percentage points (MHA$^*$) and by less than one point for four of the six
planners; pRRTC is identical problem for problem. Once the non-time caps are
removed, success is already close to saturated at 1\,s: it was removing the caps,
not lengthening the clock, that changed the picture. The one planner that
genuinely consumes the extra second is the anytime one, and it spends it on
quality rather than feasibility, with MIT$^*$'s success moving by half a point while
its mean path cost improves from 6.59 to 6.47. That is the behavior an anytime
method should exhibit, and it is legible only because the budget is equal and the
collision predicate is shared.

\paragraph{Benchmark harness.}
All numbers are produced by benchmark entry points that ship with the library; we
describe them here by what they do. The problem loader ingests
MotionBenchMaker problems (the VAMP pickles or the \texttt{robometrics} datasets),
normalizes their box/sphere/cylinder geometry into conservative OBBs, and runs
every planner on the shared backend, emitting a per-trial record and an aggregate
summary. One equal-budget driver produces every number in
\S\ref{sec:exp-singlearm}--\S\ref{sec:xparadigm}: it takes the wall-clock budget
as an explicit argument and stamps that one value onto all six planners, so an
equal-budget comparison has to be asked for rather than inherited. Asking for it
matters, because the shipped per-planner configurations are \emph{not} equal---they
hand the sampling planner 4000\,ms and the search and optimization planners
1000\,ms---and that asymmetry is exactly what the audit above is about. Every
result in this paper is therefore reported together with the budget it was
obtained under: 2\,s equal for the 800-problem cross-paradigm study
(Table~\ref{tab:singlearm}, Figs.~\ref{fig:xparadigm}--\ref{fig:xparadigm-heatmap})
and the multi-robot comparison, and the budget stated in each caption or sentence
for the fixed-subset intra-solve sweeps and ablations, where a planner is
sometimes deliberately given less or more time to show what a budget buys it. The budget actually used is recorded in the run's own summary artifact
alongside the results. Without that argument each planner falls back to its own
configured limit, which is useful for per-planner regression but is not an
equal-budget comparison. A separate
scaling driver performs the intra-solve study of \S\ref{sec:exp-throughput},
sweeping the sampling planner's parallel width, tree capacity, and edge
granularity on a fixed problem subset while recording solve time, success, and
device-memory delta per setting.

\subsection{Intra-Solve Scaling and Memory Footprint}
\label{sec:exp-throughput}
The shipped API solves one problem per call, so we characterize the parallelism
it does expose, namely parallel width within a single solve, rather than a
concurrent batch-of-$B$ throughput the code cannot yet produce
(\S\ref{sec:batched}). We sweep the sampling planner's parallel tree-extension
width $W$ and its per-tree node capacity on a fixed 24-problem MBM subset, 1\,s budget, reporting
median solve time over solved problems and the planner's own device-memory
footprint (a delta above a per-point baseline, since a co-resident process
occupies the card).

\begin{figure}[t]
  \centering
  \includegraphics[width=0.49\linewidth]{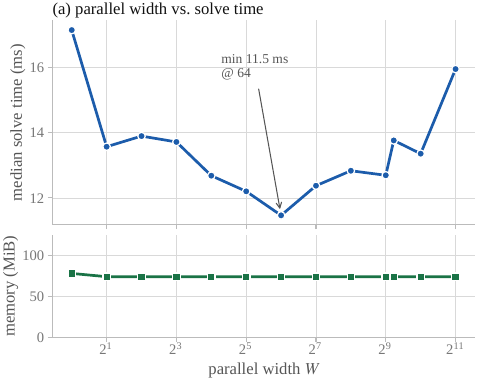}\hfill
  \includegraphics[width=0.49\linewidth]{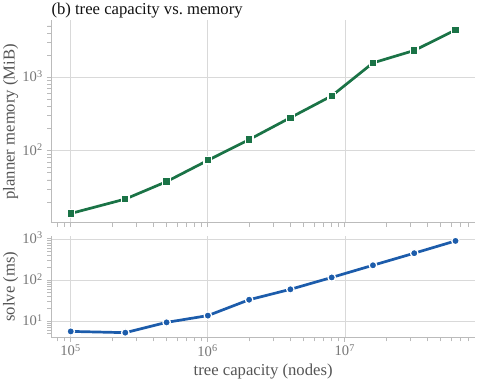}
  \caption{Intra-solve scaling of the sampling planner (pRRTC), fixed 24-problem
  MBM subset, 1\,s budget, RTX~3060. Each panel stacks its two measures on
  separate axes over a shared $x$ axis, so each is read on its own scale.
  \textbf{(a)} median solve time (top) and planner device memory (bottom) vs.\
  parallel width $W$ on a log axis: a shallow-U latency
  curve, with memory flat at $\approx$74\,MiB and hence independent of width.
  \textbf{(b)} planner memory (top) and median solve time (bottom), both on log
  axes, vs.\ tree capacity: the device-memory ceiling grows
  linearly with capacity while all settings still solve 24/24.}
  \label{fig:scaling-width}
\end{figure}

\paragraph{Parallel width.}
Figure~\ref{fig:scaling-width}(a) shows a shallow ``U'': median solve time
falls from 17.2\,ms at $W\!=\!1$ to a minimum of 11.5\,ms near $W\!=\!64$, then
rises gently to 16.0\,ms at $W\!=\!2048$ as per-iteration overhead outweighs the
extra explored width. All 24 problems solve at every width. The planner's memory
footprint is flat at $\approx$74\,MiB across the entire
$1$--$2048$ range: parallel width is essentially free in memory, because width
adds parallel blocks over the shared trees rather than enlarging them. The
practical reading is that a moderate width ($64$--$128$) is the latency sweet
spot and there is no memory penalty for choosing it.

\paragraph{Memory ceiling vs.\ tree capacity.}
Memory is instead governed by tree capacity. Figure~\ref{fig:scaling-width}(b)
sweeps tree capacity from $10^5$ to $6.4\!\times\!10^7$ nodes: the planner
footprint grows essentially linearly with capacity, from 14\,MiB at $10^5$ to
4.4\,GiB at $6.4\!\times\!10^7$, while median solve time grows in step (5.7\,ms to
0.9\,s) because a larger capacity lets the tree grow deeper before the budget or a
solution stops it. Every setting still solves 24/24, so what the sweep shows is
the shape of the memory ceiling, not a success cliff: on a 12\,GiB card the
sampling planner remains comfortable up to tens of millions of nodes, far beyond
what single-arm MBM problems require ($10^5$--$10^6$ suffices). This is the
memory-versus-capacity trade-off referenced in \S\ref{sec:discussion}.

\subsection{Single-Arm Results (7-DoF Panda)}
\label{sec:exp-singlearm}
Table~\ref{tab:singlearm} gives the numeric single-arm results underlying the
cross-paradigm figures of \S\ref{sec:xparadigm}: all six planners on the full
800-problem MBM Panda set, 2\,s equal budget, shared bit-identical backend. The
comparison is internal, across paradigms on one substrate; a head-to-head against
upstream pRRTC or OMPL is not run here (\S\ref{sec:exp-setup}), so the point is
the paradigm spread and not a speed win over a specific external build. What we
add over the compile-time-fixed upstream pRRTC, namely running an arbitrary
runtime robot on the shared backend, is architectural, and this table exercises it
twice over: every row is driven through one runtime robot-model
interface, and the same six binaries produce
Table~\ref{tab:crossrobot} for a robot with a prismatic torso and for a 14-DoF
two-arm composite without recompilation.

\begin{table}[t]
  \centering
  \caption{Single-arm results on 800 MotionBenchMaker Panda problems (7-DoF),
  2\,s equal budget, RTX~3060, shared collision/kinematics backend. Median solve
  time is over solved problems (wall-clock; no kernel/copy split in this
  release). ``Feasible'' excludes the 15 problems whose goal configuration is
  already in collision and which no planner can solve. Path cost is
  configuration-space length for sampling/search/anytime and is not
  commensurable with STOMP/CHOMP's internal objective, so the latter are omitted
  rather than mixed; it is the mean over each planner's own solved problems, so
  the harder tail a stronger planner also solves is included in its mean. The
  swatch beside each planner is the color that planner carries in every figure
  of this paper.}
  \label{tab:singlearm}
  \small
  \begin{tabular}{@{}llrrrrl@{}}
    \toprule
    Planner & Paradigm & Success & Feasible & Median (ms) & Cost & Notes \\
    \midrule
    \pmark{ppSampling}pRRTC   & sampling            & 98.1\% (785/800) & 100.0\% & 11.0  & 6.95 & all feasible solved \\
    \pmark{ppSearchLight}wPA$^*$SE & parallel search   & 94.5\% (756/800) & 96.3\%  & 3.0   & 6.82 & fastest median \\
    \pmark{ppSearch}MHA$^*$    & multi-heur.\ search & 90.2\% (722/800) & 92.0\%  & 3.4   & 7.56 & costliest paths \\
    \pmark{ppOpt}STOMP   & optimization (DF)   & 77.1\% (617/800) & 78.6\%  & 18.9  & --- & fast when it converges \\
    \pmark{ppSamplingLight}MIT$^*$    & anytime-opt.\ sampl.& 77.0\% (616/800) & 78.5\%  & 2008.8 & \textbf{6.47} & best paths, full budget \\
    \pmark{ppOptLight}CHOMP   & optimization (grad) & 10.0\% (80/800)  & 10.2\%  & 3.7   & --- & needs a seed \\
    \bottomrule
  \end{tabular}
\end{table}

\subsection{Whole-Body Composite Results (14-DoF Baxter)}
\label{sec:exp-wholebody}
Nothing in the substrate knows how many arms a robot has: it sees a sphere set
and an ACM (\S\ref{sec:wholebody}), so a whole-body composite needs no
per-paradigm re-engineering and presents only a larger $d$ and more spheres to
the same backend. We test the claim instead of assuming it. The Baxter
model is the dual-arm composite: both 7-DoF arms fused into one 14-DoF kinematic
tree with 75 spheres and an inter-arm self-collision specification of 80
sphere-range pairs, loaded at runtime through the same interface as the
7-DoF Panda and the 8-DoF Fetch. No planner was modified to
accept it, and the same six binaries produced Table~\ref{tab:crossrobot}.

\begin{table}[t]
  \centering
  \caption{Success rate over feasible problems (goal not already in
  collision) at a 2\,s equal budget, on three robots of increasing dimension
  through one unmodified substrate. Median solve time over solved problems in
  parentheses; ``---'' where a planner solved nothing. Feasible counts are
  785/800 Panda, 678/700 Fetch, 272/300 Baxter.}
  \label{tab:crossrobot}
  \small
  \begin{tabular}{@{}lrrr@{}}
    \toprule
    & Panda (7 DoF) & Fetch (8 DoF) & Baxter (14 DoF) \\
    \midrule
    \pmark{ppSampling}pRRTC     & \textbf{100.0\%} (11\,ms)  & \textbf{100.0\%} (21\,ms) & \textbf{90.1\%} (50\,ms) \\
    \pmark{ppSearchLight}wPA$^*$SE & 96.3\% (3.0\,ms) & 50.4\% (119\,ms) & 11.0\% (714\,ms) \\
    \pmark{ppSearch}MHA$^*$      & 92.0\% (3.4\,ms) & 46.3\% (136\,ms) & 32.0\% (488\,ms) \\
    \pmark{ppOpt}STOMP     & 78.6\% (19\,ms)  & 8.0\% (452\,ms)  & 16.9\% (1084\,ms) \\
    \pmark{ppSamplingLight}MIT$^*$      & 78.5\% (2009\,ms)& 35.4\% (2022\,ms)& 0.0\% (---) \\
    \pmark{ppOptLight}CHOMP     & 10.2\% (4\,ms)   & 0.0\% (---)      & 0.0\% (---) \\
    \bottomrule
  \end{tabular}
\end{table}

Three points stand out. First, the architectural claim survives
the test: a 14-DoF two-arm composite with inter-arm self-collision runs on the
same substrate with no per-paradigm work, and sampling still solves 90\,\% of the
feasible problems on it. Second, dimension separates the paradigms far more
sharply than clutter does. Between Panda and Baxter, sampling loses about ten
points while every other paradigm collapses: wPA$^*$SE falls from 96.3\,\% to
11.0\,\% and MIT$^*$ from 78.5\,\% to nothing. Third, the search ordering
inverts: wPA$^*$SE beats MHA$^*$ on both single-arm robots but is beaten by
it three-to-one on the composite (11.0\,\% vs.\ 32.0\,\%), so the choice between
the two search methods depends on dimension and is not settled.

MIT$^*$'s 0/272 is a property of the algorithm family at this dimension, not a
defect of our port, and we verified as much instead of assuming it. Three
observations separate the two explanations. First, it is not a crash or an
early exit: on every feasible problem the planner runs the full budget (median
2001\,ms) and builds a substantial graph (a median of 16\,904 vertices over 657
batches) before reporting no solution; the vertex count varies by less than
20\,\% across problems, so every problem fails the same way. Second, the
reference implementation fails identically: OMPL's CPU MITstar, run on the same
problems through a byte-identical robot model, also solves 0 of 30 sampled Baxter
problems, at a 1\,s budget and again at a full 10\,s. Third, it is not a
parameter choice. We swept every knob that governs first-solution discovery: the
RGG factor $\gamma$ over a twentyfold range (2.0 to 0.1), the batch size over a
sixtyfold range (64 to 4096), the neighbor and per-node edge limits (32 to 64),
the collision-check resolution (0.025 to 0.1\,rad), and the
estimated-initial-solution heuristic on and off. At the 2\,s equal budget every
one of these settings returns 0 of 60. That the knobs are live is
confirmed on Panda, where the same $\gamma$ sweep moves success from 78\,\% to
17\,\%. Only by breaking the equal-budget protocol does anything change at all:
at a 10\,s budget with a 4096 batch and a 400\,000-vertex graph (five times the
wall clock the other planners receive and sixty-four times the batch) MIT$^*$
solves exactly 1 of 60. Meanwhile pRRTC solves 90\,\% of the same problems through the same
collision predicate, so feasible solutions exist and the substrate admits them.
We read this as batch-then-lazily-verify graph construction being starved in 14-D
narrow passages, precisely where RRT-Connect's bidirectional greedy extension
thrives. The shared substrate is what makes the limit visible, since the
collision checker is held fixed across the comparison. The CPU
control was taken on a 30-problem sample rather than the full 300, which is the
one loose end in this argument.

\paragraph{Whole-body inverts the paradigm ordering (Gap 3).}
Section~\ref{sec:intro} posed the question this composite exists to settle: on
whole-body high-DoF systems the only GPU-native option today is trajectory
optimization, and whether sampling or search is viable at these dimensionalities
on the same backend was unknown. The measurement inverts that presumption.
Optimization, the paradigm the existing GPU tooling provides, is the weakest of
the three here: CHOMP solves nothing at all on the composite and STOMP
16.9\,\%. Sampling, whose viability was the open
question, solves 90.1\,\% of the feasible problems. Search lands in between and
splits internally. The answer to Gap 3 is therefore not that sampling and search
also work, but that the whole-body regime reverses the paradigm ordering that
single-arm intuition and the current tooling would predict. Stating
it requires the shared backend: across separate implementations, the distance
between CHOMP's 0\,\% and pRRTC's 90\,\% could always have been charged to a
difference in collision checking rather than to the paradigms.

\subsection{Cross-Paradigm Comparison at an Equal Budget}
\label{sec:xparadigm}
This is the study the shared backend enables: pRRTC, MIT$^*$, MHA$^*$,
wPA$^*$SE, STOMP, and CHOMP on identical problems, an identical scene, and an
identical collision/kinematics backend, driven through a single planning entry point. The six span all three paradigms, with search represented
by both a multi-heuristic method (MHA$^*$) and a parallel bounded-suboptimal method
(wPA$^*$SE), and optimization by both a derivative-free method (STOMP) and a
gradient-based method (CHOMP) whose obstacle gradient is the finite-difference
derivative of the same summed-hinge cost the other planners query. Because the
boolean collision predicate is bit-identical across planners
(Section~\ref{sec:backend}), differences reflect the planners rather than their
collision checkers, so we can say when each paradigm is preferable (feasibility
speed, anytime path quality, bounded suboptimality, or smooth optimization) and
read that as an empirical guide. Because wPA$^*$SE is the same algorithm SRMP
runs on the CPU, its GPU-substrate result also isolates the substrate's
contribution for a fixed algorithm.

\paragraph{Comparison protocol.}
We run all six planners on the full MotionBenchMaker Panda set (800 problems
across 8 scene families, 100 each), obtained from the \texttt{robometrics}
distribution and loaded through the shared harness, with a 2\,s per-problem time
budget on a single RTX~3060, audited so that the clock---not an iteration or
buffer cap---is what ends every planner's solve. Every planner sees the same
start/goal pairs and the same OBB scene, and queries the same bit-identical
collision predicate (Section~\ref{sec:backend}). Median solve time is reported over solved
problems only, so a planner that burns the full budget on failures is not
credited with a fast median.

\begin{figure}[t]
  \centering
  \includegraphics[width=\linewidth]{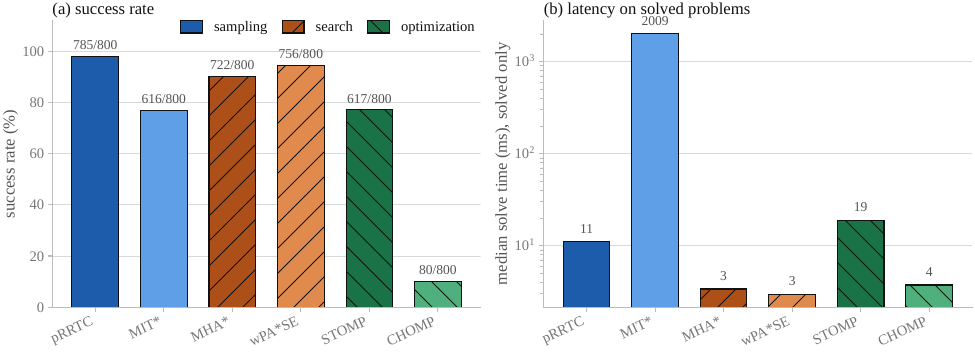}
  \caption{Cross-paradigm comparison on 800 MotionBenchMaker Panda problems,
  shared collision/kinematics backend, 2\,s equal budget. (a)~Success rate
  (solved/attempted annotated; the denominator includes the 15 problems whose
  goal is already in collision); (b)~median solve time on solved problems (log
  scale). Bar hue marks the paradigm (sampling: blue; search: orange;
  optimization: green), and within a hue the lighter shade is the second planner
  of that paradigm; the hatch angle repeats the paradigm as a second,
  color-independent channel (sampling unhatched, search forward-hatched,
  optimization back-hatched) so the grouping survives grayscale printing.}
  \label{fig:xparadigm}
\end{figure}

\begin{figure}[t]
  \centering
  \includegraphics[width=\linewidth]{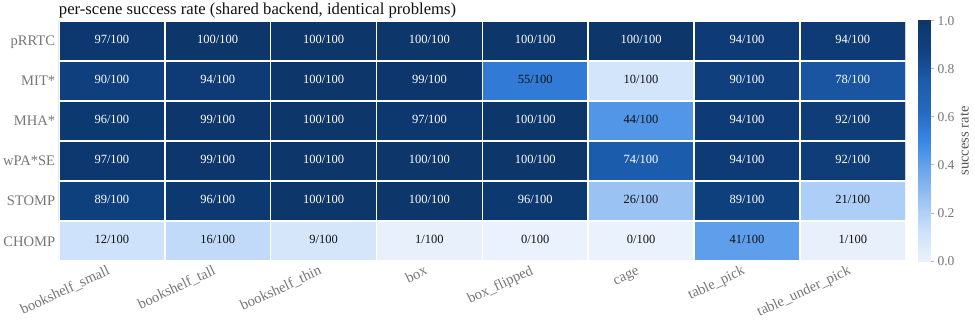}
  \caption{Per-scene success rate (planner $\times$ scene family) on the same
  800 problems and shared backend, 2\,s equal budget. At this budget the open
  \texttt{box} and \texttt{box\_flipped} scenes no longer separate the
  paradigms (every planner but CHOMP is at or near 100\%, the lone exception being
  MIT$^*$ at 55/100 on \texttt{box\_flipped}), and the discriminating
  family is \texttt{cage}, where success falls monotonically across the paradigm
  ordering (pRRTC 100, wPA$^*$SE 74, MHA$^*$ 44, STOMP 26, MIT$^*$ 10, CHOMP 0 of 100).
  \texttt{table\_under\_pick} is the second discriminator and is where the local
  optimizers collapse (STOMP 21, CHOMP 1).}
  \label{fig:xparadigm-heatmap}
\end{figure}

\paragraph{Results.}
Figure~\ref{fig:xparadigm} shows a clear feasibility ranking, but reliability and
latency do not rank the planners the same way, and that matters for how the
results should be read. Sampling (pRRTC) solves \textbf{98.1\%} (785/800), which
is every problem whose goal is not already in collision: on this set,
under this budget, it never fails a solvable problem. The two search planners
follow closely, wPA$^*$SE at \textbf{94.5\%} (756/800) and MHA$^*$ at
\textbf{90.2\%} (722/800). That 4.3-point gap sits just outside the
$\pm$2.6-point 95\,\% interval on a difference of two proportions at this sample
size, so it is real but small: the parallel bounded-suboptimal method and the
multi-heuristic one are near-equivalent on
7-DoF single-arm problems, and the case for carrying both rests on the
composite results of \S\ref{sec:exp-wholebody}, where they invert, rather than on
a gap here. The anytime-optimal sampler MIT$^*$ solves \textbf{77.0\%} while
spending its full budget refining (median 2009\,ms, the entire 2\,s), which is the
latency--quality trade-off an anytime method is supposed to make. The optimizers split:
STOMP reaches \textbf{77.1\%}, on par with MIT$^*$, while CHOMP manages
\textbf{10.0\%}.

On latency the ordering reverses. The search planners post the lowest medians on
the problems they solve, wPA$^*$SE 3.0\,ms and MHA$^*$ 3.4\,ms against pRRTC's
11.0\,ms, so sampling buys its last few points of reliability with roughly
$3\text{--}4\times$ the median latency. Because medians are taken over solved
problems only, this is a statement about the easy majority: wPA$^*$SE's mean is
84.6\,ms against a 3.0\,ms median, so a small tail of hard problems consumes most
of its budget. The per-scene view (Fig.~\ref{fig:xparadigm-heatmap}) shows
where: at a 2\,s budget the open \texttt{box} and
\texttt{box\_flipped} families are solved by everything except CHOMP---and,
on \texttt{box\_flipped}, MIT$^*$---and the
whole paradigm spread concentrates in \texttt{cage}, where success falls
monotonically from sampling (100/100) through parallel search (74/100),
multi-heuristic search (44/100) and STOMP (26/100) to MIT$^*$ (10/100) and
CHOMP (0/100).

\paragraph{Path quality, where it is commensurable.}
Cost cannot be compared across paradigms (\S\ref{sec:exp-setup}), but it can be
compared among the four planners that all report
configuration-space path length, and doing so confirms the guarantee each
algorithm advertises. MIT$^*$ returns the cheapest paths (mean 6.47), ahead of
wPA$^*$SE (6.82) and pRRTC (6.95), with MHA$^*$ the most expensive at 7.56, or
17\,\% above MIT$^*$. The anytime-optimal planner therefore does buy what it trades
latency for: MIT$^*$ spends its entire budget and returns both the fewest solutions
and the best ones. We state that as a measurement, not as a property assumed from
the algorithm's name. The ordering also puts the search
planners' suboptimality bound in perspective. wPA$^*$SE runs at a heuristic
inflation of $w\!=\!5$ and MHA$^*$ at an inflation of 5 behind an anchor factor of
2, so both bounds are satisfied by a wide margin: the worst planner is 17\,\%
above the best, not 400\,\%. The guarantee is real but loose, and we prefer to
report the measured gap rather than let the bound imply near-optimality.

\paragraph{When to use which.}
On this shared substrate the empirical guide is: use \emph{sampling} (pRRTC) when
feasibility is paramount or the robot is high-dimensional---it is the only
paradigm that stays above 90\,\% from 7 to 14 DoF---and accept a
$3\text{--}4\times$ median-latency premium for that reliability; use
\emph{parallel search} (wPA$^*$SE) when a bounded-suboptimality guarantee is
required and the robot is low-dimensional, where it is both the fastest to a
solution and within a few points of sampling on success; prefer
\emph{multi-heuristic search} (MHA$^*$) over it as dimension grows, since the two
invert on the 14-DoF composite; use \emph{anytime-optimal sampling} (MIT$^*$) when
path quality matters more than latency, a full budget is available, and the
dimension is modest; and reserve the \emph{optimizers} (STOMP/CHOMP) for refining
an already-feasible seed rather than solving from scratch in clutter. We do not
compare path cost across paradigms here: the sampling/search planners report
configuration-space path length ($\approx$7) whereas STOMP/CHOMP report their
internal objective (thousands), so the cost columns are not commensurable
(Section~\ref{sec:backend}).

\paragraph{Not every paradigm can use a budget.}
Doubling the budget is not uniformly useful, and the shared substrate makes the
difference measurable instead of anecdotal. CHOMP solves \textbf{80 of 800}
Panda problems, and solves the same 80 whether its iteration cap is 100 or
$10^5$ and whether the budget is 1\,s or 2\,s; the results are bit-identical.
That is structural, not a tuning failure: CHOMP is deterministic gradient descent
from a straight-line seed with no random restarts, so once it reaches a local
minimum further iterations cannot move it, and additional wall-clock buys nothing.
STOMP, whose derivative-free rollouts do resample, converts the same budget into
success. ``Give it more time'' is therefore a remedy for stochastic and anytime
methods and no remedy at all for deterministic local optimization, and reporting
CHOMP as the weakest planner here is a statement about that structure, not about
its implementation.

\paragraph{What the shared backend buys (Gaps 1 and 2).}
The two remaining gaps of \S\ref{sec:intro} close here, and neither closes as a
measurement. Gap 1 asked for paradigms exposed as first-class, independently
invocable modules rather than as seeding subroutines inside an optimizer: every
row of this study is a separate planner, chosen by name through one entry point,
and the six are interchangeable without touching the substrate or recompiling
(\S\ref{sec:iface}). Gap 2 asked for a comparison in which the collision
predicate and the forward kinematics are held fixed: they are the same device
routine and the same FK for all six (\S\ref{sec:backend}), which is what licenses
reading the spread from pRRTC's 98.1\,\% to CHOMP's 10.0\,\% as a statement about
algorithms. Both are properties of the architecture rather than results in their
own right. Without them the numbers above would be a comparison of
implementations, and the paradigm ordering they establish, here and on the
composite, would not be attributable to the paradigms at all.

\subsection{Backend Ablation: Edge Batch Granularity}
\label{sec:exp-ablation}
\begin{figure}[t]
  \centering
  \includegraphics[width=0.55\linewidth]{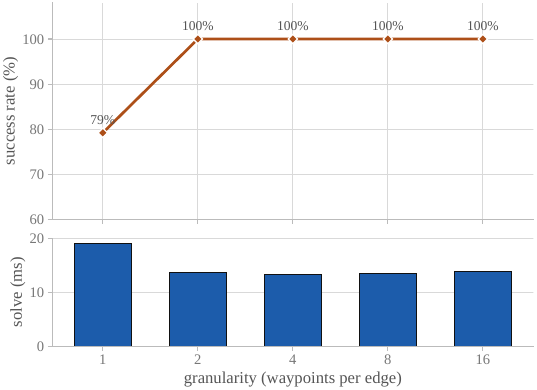}
  \caption{Edge-granularity ablation (the only runtime-toggleable backend knob),
  fixed 24-problem MBM subset, 1\,s budget. Success rate (top panel) is 24/24 for
  $g\!\ge\!2$ and collapses to 19/24 at $g\!=\!1$, where over-coarse edge checks
  admit invalid motions; median solve time (bottom panel) is flat for
  $g\!\ge\!2$.}
  \label{fig:ablation}
\end{figure}

Of the backend design choices, only one is toggleable at runtime in the shipped
code: the edge granularity $g$, the number of waypoints checked per edge (capped
at 16 by the kernel). The other three named factors, namely
the two-phase approximate$\to$detailed early exit, the structure-of-arrays layout,
and the 4-thread cooperative check, are compile-time constants in the kernel
rather than runtime switches, so we cannot ablate them without a separate build
and do not report a synthetic isolation of each. We therefore ablate $g$ alone, on the fixed 24-problem subset
(Fig.~\ref{fig:ablation}). For $g\!\ge\!2$, solve time is flat at
$\approx$13--14\,ms with 24/24 solved and memory flat at 74\,MiB: within the
useful range, denser edge checking costs almost nothing because the checks of one
edge are already batched across the cooperating threads. At $g\!=\!1$ (a single
midpoint check per edge) the planner degrades sharply: 5 of 24 problems fail and
both solve time and memory spike, because too-coarse edges admit invalid motions
that must be discovered and undone later. What the sweep yields is therefore a
practitioner default, $g$ between 2 and the cap, and not the
throughput-decomposition of the fixed backend factors that this release cannot
provide.

\subsection{Integration Surface and Demonstration Scope}
\label{sec:exp-integration}
We report no end-to-end demonstration on a physical robot or in a simulator in
this release, and no timing through the ROS\,2 path; every result in this paper
is produced by driving the six planners through the single entry point of
\S\ref{sec:iface} from the benchmark harness. The integration
surface is nonetheless concrete: alongside the Python API and the
plugin/planner abstraction of Section~\ref{sec:impl}, the release ships a ROS\,2
node that serves a native, MoveIt-free planning service and tracks the collision
world from a scene topic, so the planners drop into a ROS\,2 graph without a
MoveIt dependency. It is unit-tested at the message-translation layer, but we do
not benchmark end-to-end latency through it here. A real-robot or learning-pipeline demonstration, and
simulator adapters, remain planned; we prefer not to report capabilities the
shipped code does not exercise.


\section{Discussion and Limitations}
\label{sec:discussion}

\paragraph{When each paradigm wins.}
The cross-paradigm study (\S\ref{sec:xparadigm}) is meant to be read as a
practitioner's guide rather than a leaderboard: on a shared backend the
paradigms stop competing on collision-checker quality and start competing on
what they actually differ in. Sampling (pRRTC/cpRRTC) tends to win first-feasible
latency; anytime-optimal sampling (MIT$^*$) buys path quality when time allows;
search (MHA$^*$ and wPA$^*$SE) offers bounded suboptimality and predictable
behavior, with wPA$^*$SE's parallel expansion trading a higher heuristic weight
for lower latency; and
optimization (STOMP and CHOMP) produces smooth trajectories and handles soft
constraints where a differentiable signal exists. The three-robot study makes
this concrete and adds a dimension the single-robot version could not show.
On the 800-problem Panda set sampling leads on feasibility (pRRTC 98.1\%, every
feasible problem) but not on latency: the two search methods post lower
medians (wPA$^*$SE 94.5\% at 3.0\,ms, MHA$^*$ 90.2\% at 3.4\,ms, against pRRTC's
11\,ms), so at 7 DoF the practitioner's trade-off is a few points of reliability
against a $3\text{--}4\times$ latency premium, not a choice between a strong and
a weak paradigm. MIT$^*$ trades latency for anytime quality (77.0\% while spending
its full 2\,s), and the local optimizers split on whether they can use a budget
at all (STOMP 77.1\%, CHOMP 10.0\%). What separates the paradigms decisively is
dimension, not clutter: on the 14-DoF composite sampling holds 90.1\% of
feasible problems while parallel search falls to 11.0\%, MIT$^*$ to zero, and the
search ordering inverts. The result is best read as a mapping from problem regime,
and especially from configuration-space dimension, to paradigm, on a backend where
the collision checker is no longer a confound.

\paragraph{The shared substrate is a boolean anchor, not a total identity.}
Fair comparison rests on the boolean any-collision predicate being bit-identical
across paradigms; the optimization path additionally consumes a summed-hinge
gradient and the anytime path a min-clearance reduction. These are different
reductions of the same per-sphere geometry, not different collision
checkers: optimization's need for a differentiable signal is intrinsic to the
paradigm and not an implementation split. We therefore claim fairness at the level
of the boolean feasibility test, and state plainly that the gradient and
clearance signals are paradigm-specific by necessity.

\paragraph{The search paradigm is a CPU/GPU hybrid.}
MHA$^*$ runs its open lists and expansions on the CPU and offloads only batched
successor-edge evaluation to the GPU. It is not a pure-GPU search, and its scaling is
bounded by host-side search overhead and CPU$\leftrightarrow$GPU transfer of edge
batches, not by collision throughput alone. We report it as a hybrid and not as a
GPU search; a fully GPU-resident search frontier is future work.

\paragraph{GPU memory and scaling.}
Within a single solve the dominant device buffer is the per-tree node capacity;
parallel tree-extension width costs almost nothing. Our capacity sweep
(\S\ref{sec:exp-throughput}) shows planner memory growing linearly with
tree capacity, from 14\,MiB at $10^5$ nodes to 4.4\,GiB at $6.4\times10^7$
nodes, with median solve time rising in lockstep (5.7\,ms to 0.9\,s) and no
success cliff on the 12\,GiB card; the ceiling is set by capacity and not by any
feasibility limit we reached. Intra-solve width is a shallow U (fastest near
$W\!=\!64$) at a flat 74\,MiB footprint. All of this is intra-solve scaling. The
shipped API takes one problem at a time, so a true concurrent batch-of-$B$
throughput curve, in which per-problem trees and scene tensors would multiply and
make $B$ memory-bound, remains future work rather than a measured result.

\paragraph{Scope of the claims.}
\sysname targets whole-body composite systems, meaning multiple arms fused into a
single high-DoF kinematic tree with an allowed-collision matrix. It does not target
multi-agent coordination, where each agent plans with its own decision variables
and inter-agent constraints. Dynamic obstacles, non-holonomic and floating-base
platforms, and manifold/task constraints beyond what cpRRTC handles are likewise
out of current scope. Two backend notes bound how our numbers should be read.
First, on geometry: robot links are sphere sets, and the substrate checks them
against oriented bounding boxes, an in-house ESDF, and mesh obstacles through the
same per-sphere machinery. The MotionBenchMaker harness, however,
conservatively over-approximates every obstacle primitive to a single enclosing
OBB, so the collision predicate exercised by our reported numbers is the
sphere-vs-OBB path; the ESDF and mesh paths are implemented and exposed through
the identical substrate but are not exercised by this evaluation. Second, the
only runtime-toggleable backend knob is edge granularity (waypoints checked per
edge, $\le\!16$); the other named backend factors (two-phase early exit,
structure-of-arrays layout, and the 4-thread cooperative check) are compile-time
constants, so their contribution is a design choice we describe but do not ablate
at runtime.

\section{Conclusion}
\label{sec:conclusion}

We presented \sysname, a motion-planning library built on the observation that
sampling-, search-, and optimization-based planning share a single GPU
bottleneck, batched collision detection and batched kinematics, and can therefore
be reduced onto one shared GPU primitive substrate. The contribution is
not another multi-paradigm framework (MoveIt already unifies paradigms on the
CPU) but the substrate beneath it: a per-sphere geometry layer that all three
paradigms consume zero-copy through paradigm-specific reductions, with a
bit-identical boolean collision predicate that makes cross-paradigm comparison
well posed for the first time. On top of this substrate the library adds
runtime-arbitrary robot support and a uniform timing/result contract, with a
whole-body composite that all three paradigms consume unmodified, demonstrated
here on a 14-DoF dual-arm Baxter as well as on 7-DoF and 8-DoF single arms.
Empirically, the shared-backend study of 1800 MotionBenchMaker problems across
those three robots yields a cross-paradigm ranking that is clean but not
one-dimensional: sampling is the most reliable paradigm everywhere (pRRTC 98.1\%
on Panda, and every feasible problem on both single-arm robots) yet is
$3\text{--}4\times$ slower in median latency than the search methods that trail
it by a few points at 7 DoF (wPA$^*$SE 94.5\%, MHA$^*$ 90.2\%); the local optimizers
are fast but seed-dependent; and configuration-space dimension, not clutter,
is what ultimately separates the paradigms---at 14 DoF sampling still solves
90\,\% of feasible problems while parallel search falls to 11\,\% and
anytime-optimal sampling to zero. This turns the substrate from an architectural
claim into a measured practitioner's guide.
Because the substrate is agnostic to the planner built atop it---a new algorithm
need only phrase its queries as one of the batched reductions---the three
paradigms instantiated here are a first population of a general backend rather
than its full extent.

\paragraph{Future work.}
Several directions follow from this release. The substrate can be populated with
further planners, both additional paradigms (learning-based and hybrid methods,
say) and new algorithms within the existing three, which would test how far the
``phrase-it-as-a-batched-reduction'' recipe generalizes. The others are a fully
GPU-resident search frontier that removes MHA$^*$'s CPU open list;
dynamic-obstacle replanning on the streaming ESDF; broader robot classes
(floating-base, non-holonomic); and multi-agent multi-arm coordination, of which
the present whole-body composite is the single-decision-maker special case.

\paragraph{Availability.}
\sysname is released as a self-contained open-source package under the Apache-2.0
license at \repourl. Every experiment in this paper is
reproducible from the benchmark entry points described in
\S\ref{sec:exp-setup}: the equal-budget driver for the single-arm, whole-body and
cross-paradigm results (\S\ref{sec:exp-singlearm}--\S\ref{sec:xparadigm}) and the
scaling driver for the intra-solve scaling and memory-ceiling sweeps. Every
figure is regenerated from the per-trial and summary artifacts those runs emit.
The Panda problems are the publicly distributed \texttt{robometrics}
MotionBenchMaker set; the Fetch and Baxter problems are converted from the
upstream MotionBenchMaker MoveIt distribution by a conversion script in the
repository.

\bibliographystyle{unsrtnat}
\bibliography{references}

\end{document}